\documentclass[runningheads]{llncs}

\usepackage[T1]{fontenc}
\usepackage{graphicx,verbatim}
\usepackage{hyperref} 
\usepackage{booktabs}
\usepackage{makecell}
\usepackage[table]{xcolor}
\usepackage[export]{adjustbox}
\usepackage{color}

\usepackage[numbers]{natbib}

\usepackage{tikz}
\usepackage{amsmath}
\usepackage{amssymb}
\usetikzlibrary{shapes,positioning}

\begin{document}

\title{Learning Cardiac Motion Priors for Implicit Neural Representations}
\titlerunning{Cardiac Motion Priors for INRs}

\author{Andrew Bell \inst{1}\orcidID{0009-0005-2560-8172} \and
George Webber\inst{1}\orcidID{0009-0003-2573-7690} \and
Steffen E Petersen \inst{2}\orcidID{0000-0003-4622-5160} \and
Andrew P King \inst{1}\orcidID{0000-0002-9965-7015} \and
Muhummad Sohaib Nazir\inst{1,3}\orcidID{0000-0003-2749-1033} \and
Alistair Young \inst{1}\orcidID{0000-0001-5702-4220}
}

\authorrunning{A. Bell et al.}

\institute{School of Biomedical Engineering and Imaging Sciences, King's College London, United~Kingdom\and
William Harvey Research Institute, Queen Mary University of London, United~Kingdom \and Cardio-Oncology Service, Royal Brompton and Harefield Hospitals, London, United~Kingdom}

\maketitle      

\begin{abstract}

Implicit neural representations (INRs) are well suited to cardiac motion estimation, providing continuous, compact representations of motion fields. However, fitting an INR to each image sequence is time-consuming and sensitive to the optimisation trajectory. Learned priors can help guide optimisation towards plausible motion fields and enable faster adaptation, but learning priors for cardiac motion INRs remains under-explored. In this work, we compare four strategies for learning cardiac motion priors, including a population prior learned by joint optimisation, a consensus prior obtained by weight averaging, auto-decoders, and meta-learning. Using short-axis tagged cardiac magnetic resonance images from the UK Biobank, we evaluate their impact on tracking accuracy, motion behaviour, and adaptation trajectory. 

All learned priors substantially improved early adaptation performance compared with random initialisation. While the simple consensus prior was effective, auto-decoders recovered large deformations faster during early adaptation. Meta-learning achieved strong early performance and maintained the best adaptation trajectory over 50 iterations. The code can be found at \url{https://github.com/andrewjackbell/nvf_priors}.

\keywords{Implicit Neural Representations  \and Cardiac Motion Estimation \and Meta-learning.}

\end{abstract}

\section{Introduction}

Implicit neural representations (INRs) are continuous parametrisations of signals using neural networks. INRs have been used to represent a variety of medical data, including images, anatomical shapes, segmentation labels, and physical fields such as motion~\cite{molaei2023}. INRs are well suited to cardiac motion estimation~\cite{arratialopez2023, alvarez-florez2024, Bell2026, garzia2025}, since physiological constraints allow compact low-dimensional motion representations, despite motion being defined over high-dimensional image domains~\cite{Chandrashekara2003}.

Estimating motion from images is an ill-posed inverse problem~\cite{Fischer2008}, requiring additional constraints to recover physiologically plausible solutions~\cite{wolterink2022a}. Furthermore, per-case INR optimisation is highly time-consuming and sensitive to the optimisation trajectory. This motivates the use of learned priors~\cite{amiranashvili2024, shen2024, alvarez-florez2026} which can rapidly recover plausible solutions. However, learning priors for cardiac motion fields remains relatively unexplored, with existing works relying on simple prior formulations~\cite{alvarez-florez2026} or providing only limited analysis of the resulting motion fields~\cite{banus2026}.

In this work, we investigate learning-based approaches for constructing cardiac motion priors for INRs, including population, consensus, auto-decoder and meta-learning formulations. Using short-axis tagged cardiac magnetic resonance images from the UK Biobank, we compare these formulations to basic initialisation strategies and evaluate their effects on few-step adaptation performance and optimisation trajectory. To our knowledge, this work provides the first systematic evaluation of prior learning strategies for INR-based cardiac motion estimation.

\section{Related Work}

\subsection{Motion Estimation with INRs}

An INR can be used to approximate a velocity field $v_\theta: \mathbb{R}^{d+1} \rightarrow \mathbb{R}^d$ mapping $d$-dimensional spatial coordinate $X$ and time $t$ to a velocity vector $v_\theta(X,t)$. Since ground truth motion is not typically available, previous works optimised network parameters $\theta$ according to inter-frame correspondence~\cite{wolterink2022a, garzia2025, lowes2025}. This assumes that an image frame $I_t$ should be similar to some reference image $I_r$ after warping according to the motion derived from $v(t)$. Thus, INR parameters $\theta$ can be learnt as

\begin{equation}
\theta^\star
=
\arg\min_{\theta}
\mathcal{L}_{sim}
\bigl(
I_t,
I_r \circ v_\theta(t)
\bigr),
\label{eq:sim}
\end{equation}
where $\mathcal{L}_{sim}$ is an image similarity loss and $\circ$ denotes image warping. This objective does not uniquely identify a solution for $\theta^\star$, making the optimisation problem ill-posed. Moreover, in real imaging data, noise, artefacts, and violations of brightness constancy create a mismatch between the image similarity objective and the true underlying motion field. Regularisation terms based on smoothness or mechanics help constrain solutions to plausible motion fields~\cite{wolterink2022a}, but balancing these terms against image observations remains challenging. Furthermore, fitting a new INR for every image sequence requires an expensive per-case optimisation procedure, which can become prohibitive for high-dimensional cardiac image data. This motivates learning a prior to improve the INR initialisation.

\subsection{Learned Priors for Motion INRs}

Early works fit INRs to cardiac motion from a random initialisation, using regularisation to encourage plausible solutions~\cite{arratialopez2023,shen2024}. Alvarez-Florez et al.~\cite{alvarez-florez2024} proposed reusing parameters between successive motion fields, which accelerated fitting, but did not leverage population-level motion information. More recently, they learned a `consensus' prior by averaging weights across INRs fit to a training set. This led to faster convergence in new cases, with improved motion plausibility~\cite{alvarez-florez2026}. These results show the benefit of a learned motion prior, but weight averaging is a simple heuristic that does not explicitly model the distribution of cardiac motion.

More principled approaches include auto-decoders~\cite{park2019a}, which use latent codes to represent inter-subject variation within a distribution, and meta-learning~\cite{sitzmann2020a}, which seeks an optimal initialisation for rapid adaptation to new signals. To our knowledge, meta-learning has not been investigated for INR-based cardiac motion priors, and no systematic comparison of these strategies has been reported.

\section{Methods}

\subsection{Neural Velocity Field}

We represent cardiac motion using a neural velocity field (NVF) based on previous work~\cite{garzia2025}. The NVF is an INR which takes spatial coordinates and time as input, and predicts an instantaneous spatial velocity vector. Formally, we represent $d$-dimensional cardiac motion as an NVF $v:\mathbb{R}^{d+1} \rightarrow \mathbb{R}^d$, and obtain Lagrangian motion $\varphi_{t_0 \rightarrow t}$ of coordinate $X$ by integrating the ordinary differential equation 
\[
\frac{d \varphi_{t_0 \rightarrow t}(X)}{dt}
=
v(\varphi_{t_0 \rightarrow t}(X),t),
\quad
\varphi_{t_0 \rightarrow t_0}(X)=X,
\]
using a second-order Runge-Kutta method with three sub-steps per frame interval. The resulting trajectories define the displacement field between any pair of time points. To encourage periodicity of cardiac motion, we map each time point to a periodic encoding
$$\tau(t)=(\sin(\theta), \cos(\theta)) \quad \theta=\left(\frac{2\pi t}{T}\right),$$ 
where T is the total number of time frames~\cite{garzia2025}.

\subsection{Objective}

To optimise the NVF $v$ over an image sequence $I \in \mathbb{R}^{T\times H \times W}$ we assume in-plane appearance-correspondence and use a similarity objective as in Equation~\ref{eq:sim}. The displacement fields required for image warping are obtained by integrating the NVF between adjacent time points. Specifically, we use normalised cross-correlation (NCC) as the similarity loss between the current frame and forward- and backward-warped adjacent frames:
\[
\mathcal{L}_{\mathrm{sim}}
=
-\mathrm{NCC}
\left(
I_t,
I_{t+1} \circ \varphi_{t+1 \rightarrow t}
\right)
-
\mathrm{NCC}
\left(
I_t,
I_{t-1} \circ \varphi_{t-1 \rightarrow t}
\right),
\]
We also assume that cardiac motion is spatially and temporally smooth, and define a smoothing loss
\[
\mathcal{L}_{\mathrm{smooth}}
=
\|\nabla_X v\|^2_F
+
\left\|
\frac{\partial v}{\partial t}
\right\|_2^2.
\]
A further assumption is that the myocardium is approximately incompressible, and that the underlying motion field preserves local topology. Computing deformation gradients from integrated trajectories at each step is computationally expensive. We therefore employ a first-order approximation based on the instantaneous velocity Jacobian $\tilde{F} = \mathbb{I} + \Delta t \nabla_X v$, where $\mathbb{I}$ is the identity matrix and $\Delta t$ is the temporal interval between adjacent frames. We then encourage volume and topology preservation with respective penalties:
\[
\mathcal{L}_{\mathrm{volume}}
=
(
\det(\tilde{F})-1
)^2,
\quad 
\mathcal{L}_{\mathrm{fold}}
=
\mathrm{ReLU}(
-\det (\tilde{F}))^2,
\]The final objective $\mathcal{L}$ is given by the weighted sum of the loss terms according to hyperparameters $\lambda = \{\lambda_{\mathrm{sim}}, \lambda_\mathrm{{smooth}}, \lambda_{\mathrm{volume}}, \lambda_{\mathrm{fold}}\}$.

\subsection{Learned Prior Formulations}
\label{section:priors}

In this section, we abstract the NVF objective as a function and describe the formulation of each prior in this notation. Let $\mathcal{F}$ denote a distribution of functions and let $T=\{f_i^\star\}_{i=1}^{N}$ be a training set sampled from $\mathcal{F}$. Each function $f_i^\star:X\rightarrow Y$ is observed at coordinates $\{x_k\}_{k=1}^{K}\subset X$, and reconstruction quality is measured using an objective function $\mathcal{L}:Y\times Y\rightarrow\mathbb{R}$. The goal is to learn prior parameters that enable rapid adaptation to unseen functions sampled from $\mathcal{F}$. 

\textbf{Population Prior}
As a simple baseline, we define a population prior $\theta^P$ as the parameters obtained by jointly optimising a single INR across all training instances:
\begin{equation*}
\theta^P =
\arg\min_{\theta}
\mathbb{E}_{f^\star \sim F}
\left[
\sum_{k=1}^{K}
\mathcal{L}
\bigl(
f_{\theta}(x_k),
f^\star(x_k)
\bigr)
\right].
\label{eq:population}
\end{equation*}

\textbf{Consensus Prior}
We define a consensus prior $\theta^C$ as the mean of independently optimised parameters for each training function:
\begin{equation*}
\theta^C =
\frac{1}{N}
\sum_{i=1}^{N}
\arg\min_{\theta}
\sum_{k=1}^{K}
\mathcal{L}
\bigl(
f_{\theta}(x_k),
f_i^\star(x_k)
\bigr).
\label{eq:consensus}
\end{equation*}
Although parameter alignment is not guaranteed across independently trained INRs, consensus priors can nevertheless be meaningful when training starts from a common initialisation~\cite{alvarez-florez2026}.

\textbf{Auto-decoder Prior} Each training instance is assigned a latent code $z_i \in \mathbb{R}^{\mathcal{Z}}$. For each hidden layer $l$, a linear modulation network $m_l$ maps the latent code to layer-wise scale and shift parameters, which modulate the periodic MLP activations ~\cite{mehta2021}. Network parameters $\theta$, modulation parameters $\kappa$, and subject-latent matrix $z\in\mathbb{R}^{N\times \mathcal{Z}}$ are jointly optimised:

\begin{equation*}
(\theta^\star,z^\star, \kappa^\star)=
\arg\min_{\theta,z, \kappa}
\sum_{f_i^\star \in T}
\sum_{k=1}^{K}
\mathcal{L}
\bigl(
f_{\theta,z_i,\kappa}(x_k),
f_i^\star(x_k)
\bigr)+
\lambda_z \|z_i\|_2^2,
\label{eq:autodecoder}
\end{equation*}
where \(\lambda_z\) is a latent-space regularisation weight. During adaptation to an unseen case, the decoder and modulation parameters are fixed and a new latent code is optimised.

\textbf{Meta-learned Prior}
We use second-order Meta-SGD to jointly learn an initialisation $\theta_0$ and layer-wise adaptation step sizes $\gamma=\{\gamma_l\}_{l=1}^{L}$. Let $\theta^\epsilon$ denote the parameters obtained after adapting $\theta^*$ to a function $f^\star\sim\mathcal{F}$ for $\epsilon$ inner-loop adaptation steps. The objective is to minimise the post-adaptation loss:

\begin{equation*}
(\theta^*, \gamma)=
\arg\min_{(\theta^*, \gamma)}
\mathbb{E}_{f^\star\sim\mathcal{F}}
\left[
\sum_{k=1}^{K}
\mathcal{L}
\bigl(
f_{\theta^\epsilon}(x_k),
f^\star(x_k)
\bigr)
\right].
\label{eq:meta}
\end{equation*}
During adaptation, parameters are updated according to
\begin{equation*}
\theta^{(s+1)}_l=
\theta^{(s)}_l -
\gamma_l
\nabla_{\theta^{(s)}_l}
\mathcal{L}
\bigl(
f_{\theta^{(s)}},
f^\star
\bigr),
\qquad l=1,\ldots,L,
\label{eq:metasgd}
\end{equation*}
where $s$ is the adaptation step and $\gamma_l$ is the learnable step size associated with layer $l$.

\section{Experiments}

\subsection{Data}

The UK Biobank (UKB) is a population-based dataset with health, genomic, and imaging data. A subset of participants underwent cardiac magnetic resonance (CMR) imaging, details of which are given in~\cite{petersen2016}. In this work, we use short-axis tagged CMR, including apical, mid-ventricular and basal slices, each having 20 time frames. Reference point tracks and segmentations of the left ventricle (LV) myocardium for tagged CMR slices were obtained using semi-automated analysis (CIM Tag2D, University of Auckland) as described previously~\cite{Bell2026, young1995a}. Images and annotations were cropped to the left ventricle with 20 pixels of padding and rescaled to consistent dimensions. In this work, we sample slices randomly from existing data splits~\cite{Bell2026} which contain no subject overlap. From the respective sets we sample 500 training slices (469 subjects), 100 validation slices (98 subjects), and 100 test slices (93 subjects).

\subsection{Implementation details}

All priors initialised a SIREN network with three hidden layers of 256 nodes~\cite{sitzmann2020}. During training and adaptation, 512 random spatial coordinates were sampled per frame, yielding $K = 10{,}240$ spatio-temporal coordinates per iteration used to fit the NVF. For meta-learning, an additional 2048 query coordinates per frame were sampled to evaluate post-adaptation performance. The NVF regularisation weights were set to $\lambda_{\mathrm{smooth}}=0.01$, $\lambda_{\mathrm{volume}}=10^{-3}$, and $\lambda_{\mathrm{fold}}=10^{-3}$; the latent regularisation weight was $\lambda_z=10^{-4}$ for the auto-decoder. Adaptation learning rates were tuned on the validation set to minimise the objective after five adaptation steps, resulting in learning rates of $10^{-4}$ (population), $10^{-5}$ (consensus), and $3 \times 10^{-2}$ (auto-decoder). Meta-learning was trained using $\epsilon = 5$ inner-loop adaptation steps to match the evaluation protocol, and learned its own layer-wise learning rates. To investigate the effect of latent dimensionality, the auto-decoder was trained with various latent dimensions $\mathcal{Z}\in\{32,\allowbreak 64,\allowbreak 128,\allowbreak 256,\allowbreak 512,\allowbreak 1024\}$. 
\subsection{Evaluation}

Priors were evaluated by fitting an NVF to each test case, initialised from the corresponding prior. We compared performance after 5 iterations and the longer-term adaptation trajectory over 50 iterations. Performance was assessed using tracking accuracy, segmentation overlap, and mechanical plausibility. Specifically, the fitted NVFs were used to warp a set of reference points across the cardiac cycle, which were compared to manually tracked points using mean Euclidean distance (displacement error). Furthermore, the reference segmentation mask was warped according to the fitted NVFs and compared to manually segmented time-varying masks with Dice score. Mechanical plausibility was assessed using the mean of deformation Jacobian determinants and the mean Green--Lagrange strains. These are averaged over all points in the left ventricular myocardium, and strains are rotated to circumferential and radial directions for clinical interpretation.   

\section{Results and Discussion}

Implicit neural representations are often limited by the lengthy iterative process required to fit data. Learning a prior from training data can overcome this limitation by initialising the neural network with parameters that allow fast adaptation to new data. Furthermore, in an ill-posed problem such as motion reconstruction, the optimisation may converge to different solutions depending on the chosen initialisation. For this reason, a prior can recover motion with higher accuracy than a random initialisation, even over long adaptation horizons.

\begin{table}
    \centering
    \setlength{\tabcolsep}{8pt}
        \caption{Motion estimation results after \textbf{five} adaptation steps. Best values are shaded.}
            \label{tab:results}

    \begin{tabular}{lrrrr}
\toprule
Prior  & \makecell{Error $\downarrow$\\(mm)} & \makecell{ES Error $\downarrow$ \\(mm)} & \makecell{Dice $\uparrow$ \\(LVM)}  & $\det(F)$ \\\midrule
Random & 3.45     $\pm$ 0.51      & 4.62        $\pm$ 0.77         & 0.850 & 0.99              \\
Population & 1.85     $\pm$ 0.49      & 2.69        $\pm$ 0.97         & 0.911 & 0.96             \\
Consensus  & 1.82     $\pm$ 0.49      & 2.62        $\pm$ 0.89         & 0.916 & 0.97                \\
Auto-decoder & 1.80     $\pm$ 0.48  & \cellcolor{gray!25} 2.11 $\pm$ 0.74   & 0.917 & 0.94                 \\
Meta-learning   & \cellcolor{gray!25} 1.66     $\pm$ 0.42      & 2.32        $\pm$ 0.75         &\cellcolor{gray!25} 0.921 & 0.97 \\
\bottomrule
\end{tabular}
\end{table}

\begin{table}
    \centering
    \setlength{\tabcolsep}{8pt}
        \caption{Motion estimation results after \textbf{50} adaptation steps. Best values are shaded.}
            \label{tab:results-50}

\begin{tabular}{lrrrr}
\toprule
Prior  & \makecell{Error $\downarrow$\\(mm)} & \makecell{ES Error $\downarrow$ \\(mm)} & \makecell{Dice $\uparrow$ \\(LVM)}  & $\det(F)$\\ 
\midrule
Random       & $1.61 \pm 0.38$ & $2.11 \pm 0.74$ & 0.922 & 0.98 \\
Population   & $1.63 \pm 0.41$ & $1.94 \pm 0.72$ & 0.919 & 0.96 \\
Consensus    & $1.47 \pm 0.37$ & $1.92 \pm 0.69$ & 0.927 & 0.98 \\
Auto-decoder & $1.72 \pm 0.41$ & $2.21 \pm 0.68$ & 0.918 & 0.95 \\
Meta-learning& \cellcolor{gray!25} $1.43 \pm 0.35$ & \cellcolor{gray!25} $1.86 \pm 0.65$ & \cellcolor{gray!25} 0.928 & 0.97 \\
\bottomrule
\end{tabular}

\end{table}

Table \ref{tab:results} shows that all learned priors substantially outperformed random initialisation after five adaptation steps, reducing the mean displacement error from 3.45 mm to 1.66--1.85 mm. Meta-learning achieved the lowest overall displacement error (1.66 mm), whereas the auto-decoder prior had the lowest end-systolic displacement error (2.11 mm). All methods produced deformation fields with mean Jacobian determinants close to 1 in the LV and no cases of folding. While metrics between prior formulations were relatively similar after five adaptation steps, the differences became more apparent over longer adaptation trajectories.

\begin{figure}
    \centering
    \includegraphics[width=0.85\linewidth]{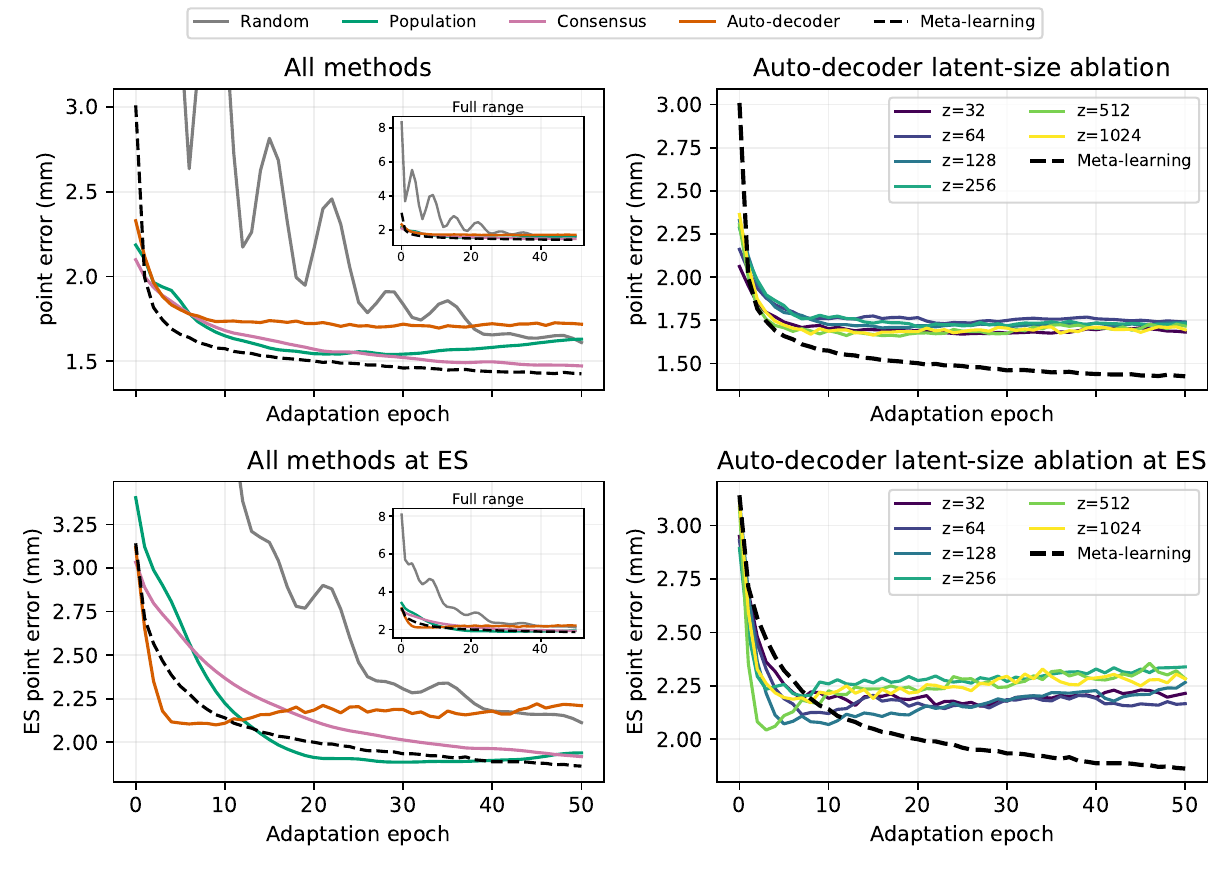}
    \caption{Adaptation trajectory of each prior (left) and various latent dimensions for auto-decoders (right). Measured is displacement error (top) and ES (end-systolic) displacement error (bottom).}
    \label{fig:trajectories}
\end{figure}

\begin{table}
\centering
\caption{Training time of each prior on 500 slices using an RTX 4090 GPU}
\label{tab:training-time}
\begin{tabular}{lr}
\toprule
Prior & Training time (min) \\
\midrule
Random Init & 0 \\
Population & 3.3   \\
Consensus & 19.3  \\
Auto-decoder & 53.4  \\
Meta-learning & 116.4 \\
\bottomrule
\end{tabular}
\end{table}

Although the meta-learned prior was optimised for five-step adaptation, performance continued to improve well beyond this horizon (Fig.~\ref{fig:trajectories}). In contrast, the auto-decoder prior converged rapidly but plateaued after approximately ten iterations. Changing the latent dimensionality had little effect on the adaptation trajectory (Fig.~\ref{fig:trajectories}, right panels), suggesting that the plateau arises from the modulation mechanism rather than the latent-space capacity. While both simple priors had good early performance, the population prior started to deteriorate after 30--40 iterations while the consensus prior continued to improve. Since the population prior is optimised to represent an average motion field, it may represent a compromise solution from which further adaptation is more difficult. Table \ref{tab:results-50} shows the full results after 50 iterations, revealing that meta-learning had the best performance across all metrics for this longer budget. Unlike the other priors, meta-learning explicitly optimises parameters for efficient adaptation to new subjects, which likely explains its superior performance. However, the chosen formulation used second-order gradients, which incurred a greater computational cost and training time relative to the other priors (see Table~\ref{tab:training-time}). 

\begin{figure}
    \centering
    \includegraphics[width=0.85\linewidth]{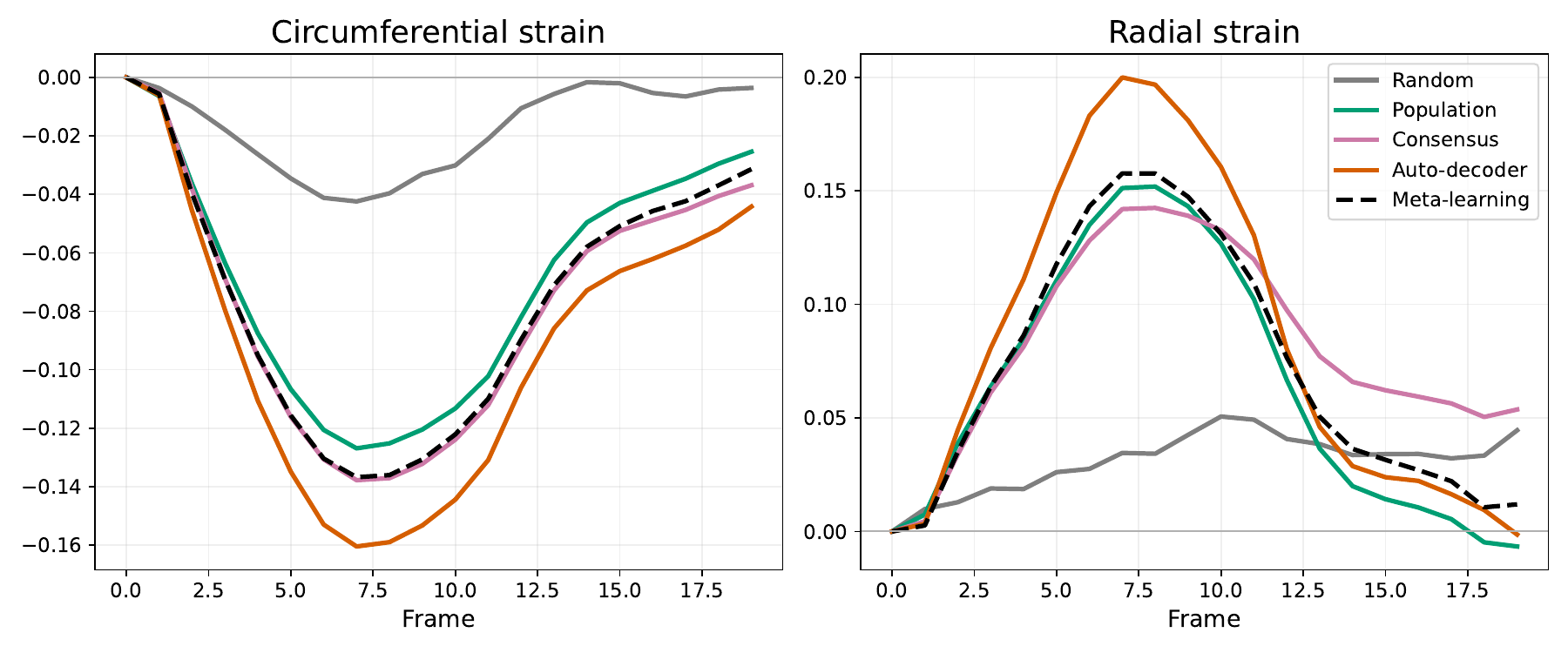}
    \caption{Circumferential (left) and radial (right) mean LV strain curves of each prior after \textbf{five} adaptation steps.}
    \label{fig:strain}
\end{figure}

\begin{figure}
    \centering
    \begingroup
    \centering
    \setlength{\tabcolsep}{2pt}
    
    \begin{tabular}{@{}lccccc@{}}
        & &  End-Diastole & End-systole & Manual GT* & \\
    
      & \multicolumn{5}{c}{%
        \includegraphics[width=0.16\linewidth]{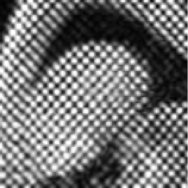}
        \includegraphics[width=0.16\linewidth]{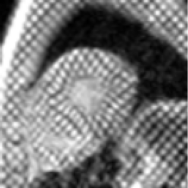}
        \includegraphics[width=0.16\linewidth]{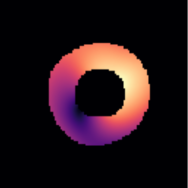}%
      }
      \\[-1pt]

      & \multicolumn{5}{c}{%
        \includegraphics[width=0.72\linewidth]{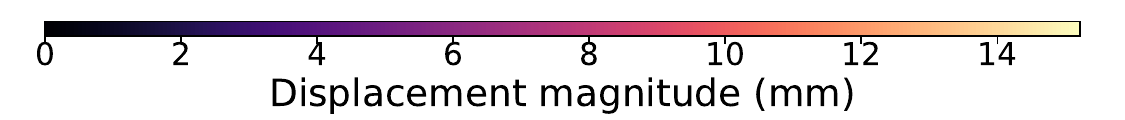}%
      }\\
    
       & Random & Population & Consensus & \makecell{Auto\\ Decoder} & \makecell{Meta\\ Learning}\\
    
      Prior & \adjustbox{valign=c}{\includegraphics[width=0.16\linewidth]{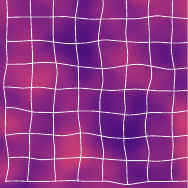}} &
      \adjustbox{valign=c}{\includegraphics[width=0.16\linewidth]{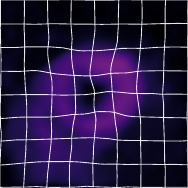}} &
      \adjustbox{valign=c}{\includegraphics[width=0.16\linewidth]{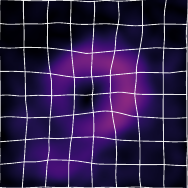}} &
      \adjustbox{valign=c}{\includegraphics[width=0.16\linewidth]{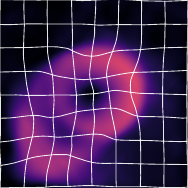}} &
      \adjustbox{valign=c}{\includegraphics[width=0.16\linewidth]{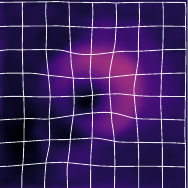}}
      \\[-1pt]
    
      $\downarrow$ \textit{adapt} & 15.46 & 8.72 & 8.03 & 7.58 & 7.13\\
    
      5 steps & \adjustbox{valign=c}{\includegraphics[width=0.16\linewidth]{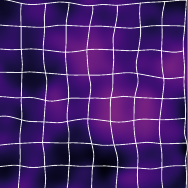}}&
      \adjustbox{valign=c}{\includegraphics[width=0.16\linewidth]{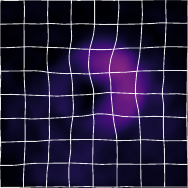}} &
      \adjustbox{valign=c}{\includegraphics[width=0.16\linewidth]{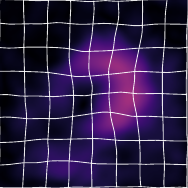}} &
      \adjustbox{valign=c}{\includegraphics[width=0.16\linewidth]{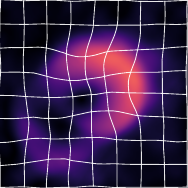}} &
      \adjustbox{valign=c}{\includegraphics[width=0.16\linewidth]{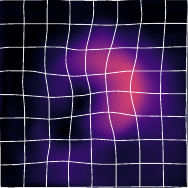}}
      \\[-1pt]
    
       $\downarrow$ \textit{adapt} & 7.00 & 7.79 & 6.94 & \cellcolor{gray!25} 5.09 & 5.64 \\

      50 steps & \adjustbox{valign=c}{\includegraphics[width=0.16\linewidth]{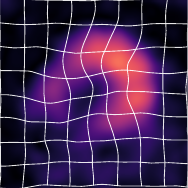}} &
      \adjustbox{valign=c}{\includegraphics[width=0.16\linewidth]{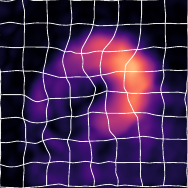}} &
      \adjustbox{valign=c}{\includegraphics[width=0.16\linewidth]{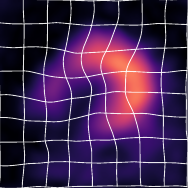}} &
      \adjustbox{valign=c}{\includegraphics[width=0.16\linewidth]{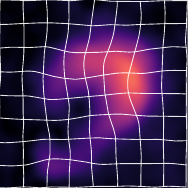}} &
      \adjustbox{valign=c}{\includegraphics[width=0.16\linewidth]{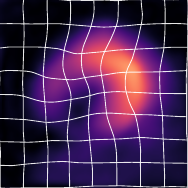}}
      \\[-1pt]
    
      & 3.66 & 3.82 & 3.49 & 4.21 & \cellcolor{gray!25} 3.07\\

    \end{tabular}
\endgroup
\caption{End-systolic displacement magnitude of each prior, adapting to a mid-ventricular test case for 0, 5 and 50 steps. The numbers underneath each image represent the error against manual ground truth (GT) in mm. *Interpolated from sparse points. Images reproduced by kind permission of the UK Biobank~© }
\label{fig:disp_maps}
\end{figure}

Fig.~\ref{fig:strain} shows that all learned priors recovered smooth physiological strain patterns after five adaptation steps. However, the auto-decoder produced larger strain magnitudes than the other methods, particularly near end-systole, consistent with its lower end-systolic displacement error. Fig.~\ref{fig:disp_maps} further shows that the auto-decoder recovers large deformations more rapidly than the other priors but exhibits little improvement with prolonged adaptation. In contrast, the population prior produces noticeably noisier displacement fields after 50 adaptation steps compared with a random initialisation. Although the population prior provides a strong starting point, these results indicate that it becomes less favourable for longer-term optimisation.

This study has limitations that should be considered when interpreting the results. Firstly, the motion estimation task was limited to 2D tagged CMR images, which have modality-specific image properties and artefacts. Secondly, due to computational limitations, the models were adapted for fixed iteration counts to assess early and longer-term adaptation performance. Therefore, fully converged performance was not evaluated in this study. Finally, the UK Biobank consists primarily of healthy individuals, limiting the diversity of motion represented in the learned priors. Priors learned from more diverse disease cohorts may exhibit different adaptation behaviour.

\section{Conclusion}

Learned priors substantially improved the speed and quality of INR-based cardiac motion estimation. A consensus prior, created by averaging INR weights, was a simple and effective initialisation. Auto-decoders were able to recover large deformations with just five adaptation steps, while meta-learning had the strongest optimisation trajectory over a 50-step adaptation horizon. These results show that the choice of learned prior dictates both adaptation speed and optimisation trajectory.

\begin{credits}
\subsubsection{\ackname} This work was supported by the Engineering and Physical Sciences Research Council (EPSRC) Doctoral Training Partnership [EP/W524475/1] and EPSRC project grant [Z5336762]. This study was conducted using the UK Biobank resource under access application 2964. 

\subsubsection{\discintname}
The authors have no competing interests to declare that are
relevant to the content of this article.
\end{credits}
\bibliography{art.bib}

\bibliographystyle{splncs04}

\end{document}